\documentclass[lettersize,journal]{IEEEtran}
\IEEEoverridecommandlockouts
\PassOptionsToPackage{hyphens}{url}\usepackage{hyperref}

\usepackage{pgfplots}
\pgfplotsset{compat=1.17}

\usepackage{caption}
\usepackage{subcaption}
\usepackage{multirow}
\usepackage{graphicx}
\usepackage{comment}
\usepackage{multirow}
\usepackage{multicol}
\usepackage{array}
\usepackage{cite}
\usepackage{amsmath,amssymb,amsfonts}
\usepackage{textcomp}
\usepackage{xcolor}
\usepackage{amsthm}
\usepackage{ulem}
\usepackage{algorithm}
\usepackage{algpseudocode}
\def\BibTeX{{\rm B\kern-.05em{\sc i\kern-.025em b}\kern-.08em
    T\kern-.1667em\lower.7ex\hbox{E}\kern-.125emX}}

\theoremstyle{definition}

\usepackage{array}
\newcolumntype{P}[1]{>{\centering\arraybackslash}p{#1}}
\makeatletter
\newcommand{\linebreakand}{
  \end{@IEEEauthorhalign}
  \hfill\mbox{}\par
  \mbox{}\hfill\begin{@IEEEauthorhalign}
}
\makeatother

\begin{document}

\title{Optimizing Control Strategies for Wheeled Mobile Robots Using Fuzzy Type I and II Controllers and Parallel Distributed Compensation
}




\author{Nasim~Paykari, Razieh~Jokar, Ali~Alfatemi,
Damian~Lyons,~\IEEEmembership{Senior Member,~IEEE,}
Mohamed~Rahouti,~\IEEEmembership{Member,~IEEE} 
\thanks{Manuscript received xx, 2023; revised xx, 2023. ({\it Corresponding author: Mohamed Rahouti}.)}
\thanks{N. Paykari, A. Alfatemi, D. Lyons, and M. Rahouti are with the Department of Computer and Information Science, Fordham University, Bronx, NY, 10458 USA (e-mail: npaykari@fordham.edu; raz.jokar24@gmail.com; aalfatemi@fordham.edu; dlyons@fordham.edu; mrahouti@fordham.edu).}
}


\maketitle

\begin{abstract}
Adjusting the control actions of a wheeled robot to eliminate oscillations and ensure smoother motion is critical in applications requiring accurate and soft movements. Fuzzy controllers enable a robot to operate smoothly while accounting for uncertainties in the system. This work uses fuzzy theories and parallel distributed compensation to establish a robust controller for wheeled mobile robots. The use of fuzzy logic type I and type II controllers are covered in the study, and their performance is compared with a PID controller. Experimental results demonstrate that fuzzy logic type II outperforms type I and the classic controller. Further, we deploy parallel distributed compensation, sector of nonlinearity, and local approximation strategy in our design. These strategies help analyze the stability of each rule of the fuzzy controller separately and map the if-then rules of the fuzzy box into parallel distributed compensation using Linear Matrix Inequalities (LMI) analysis. Also, they help manage the uncertainty flow in the equations that exist in the kinematic model of a robot. Last, we propose a Bezier curve to represent the different pathways for the wheeled mobile robot.
\end{abstract}

\text {Keywords—}
Fuzzy Type I, Type II, WMR, PDC, Takagi-Sugeno, Nonlinear, Local approximation

\section{Introduction} Mobile robots have undergone significant progress over the years and are now utilized in various applications, including manufacturing, medical robotics, and autonomous vehicles \cite{zeng2022development, bernardo2022survey, lyons2022visual}. The capacity of mobile robots to move smoothly and precisely is a crucial component, especially for wheeled mobile robots. The design and implementation of these robots' control systems significantly influence how well they perform. Wheeled mobile robots can be controlled today using fuzzy logic-based controllers, demonstrating promising results \cite{paykari2013design}. The control problem can become more complicated due to system uncertainties and interactions between various parameters \cite{simon2023fuzzy}. In this case, a practical and knowledgeable controller for wheeled mobile robots can be developed by fusing fuzzy logic theories with parallel distributed compensation. Such a controller can ensure safety, effectiveness, and overall performance in various applications. 

\begin{figure}[h]
    \centering
    \includegraphics[width=5cm]{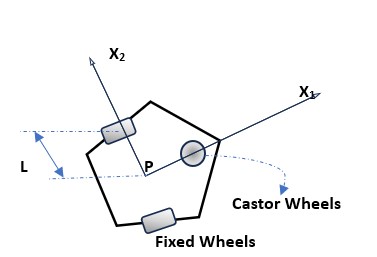}
    \caption{WMR type (2,0).}
    \label{fig:Type_2_0}
\end{figure}

We offer a novel approach that combines fuzzy logic type I and type II with parallel distributed compensation to enable a reliable and effective controller for wheeled mobile robots. Specifically, we propose to fuse Fuzzy theories \cite{wolkenhauer1997course, lopez2019review}, and Parallel Distributed Compensation \cite{wang1996approach}. We aim to provide a practical and knowledgeable controller for a wheeled mobile robot shown in Fig. \ref{fig:Type_2_0}. Providing smooth and accurate movement in a wheeled robot is critical in various applications, including precision manufacturing, medical robotics, and autonomous vehicles, as it may assure safety, efficiency, and overall performance \cite{zhang2023automated}. In precise (i.e., non-fuzzy) mathematics, the value of the set membership function for every element is either 0 or 1, whereas in fuzzy type I systems, this changes to a function with a value between 0 and 1. Moreover, Fuzzy type II systems go beyond type I; the degree of membership for a point can be a function. In other words, the membership functions of type I are of two dimensions, while the membership functions of type II are of three dimensions. The Fuzzy type I approach integrates system uncertainties into the controller equation, and the Fuzzy type II approach adds uncertainties to Fuzzy type I.

Prior studies have shown that wheeled mobile robots can be split into five groups based on mobility and steerability \cite{tagliavini2022wheeled}, as presented in Table \ref{tab:Tabel_1}. We choose the model type (2,0), for our study where the two numbers refer, respectively, to the degree of steerability and the degree of mobility \cite{campion2011structural, parhi2011kinematic}. A regular car, for example, is a (2,0) robot because it has two steerable wheels in the front and fixed wheels in the back, offering modest steerability. Lower steerability refers to a wheeled robot's ability to turn or change direction in restricted places. In contrast, higher steerability refers to the robot's capacity to perform sharper turns and travel more readily in limited locations \cite{leong2022four}. A car, for example, has less steerability than a tank, which can turn on the spot due to its differential steering mechanism. This model has the lowest motor numbers, which leads to fewer control parameters. As a result, there are fewer control parameters. Additionally, an overly complicated mathematical model will be avoided despite the potential for this, given the dimensions increase when we later employ Fuzzy type II. Further, this type of wheeled mobile robot includes a generally useful kinematic structure that all classes can share \cite{khan2022speed}. Last, with two motors, the robot's advantage of being quick and small can be implemented.

\begin{table}[h]
    \caption{Wheeled mobile robot types \cite{tagliavini2022wheeled}.}
    \label{tab:Tabel_1}
    \centering
    \begin{tabular}{c|c|c|c}
    \hline
        Type &  Steerability & 	Mobility & 	No. of Motors\\
        \hline
        (3,0)&	3&	0&	3\\
        (2,1)&	2&	1&	3\\
        (1,2)&	1&	2&	3\\
        (2,0)&	2&	0&	2\\
        (1,1)&	1&	1&	2\\
        (0,0)&	0&	0&	0\\ 
        \hline
    \end{tabular}
\end{table}

The design of a fuzzy controller needs additional human expertise beyond the kinematic equations of the model \cite{aftab2022fuzzy}. In contrast, the extension of these equations plays a more critical role in analyzing controllers and exploring the steady state. 
Extending the mathematical equations is required for stability analysis and exploration. In our work, we need to invoke the kinematic model equations of the wheeled mobile robot to have a stable controller.



When there are more uncertainty parameters in the system, i.e., when one uncertainty affects another, or a parameter interacts with two or more uncertainties, an approach leveraging Fuzzy Type II is more robust and produces better results. Our approach relies on fuzzy logic type I, which introduces uncertainty, and fuzzy logic type II, which includes type I's uncertainty. The if-then box is a notable example of how complexity in type II grows rapidly. Therefore, we propose to use parallel distributed compensation, sector of nonlinearity, and a local approximation approach to control the flow of the uncertainties in the equations. By allowing for separate analysis of each rule of the fuzzy controller and mapping the if-then rules of the fuzzy box into parallel distributed compensation using Linear Matrix Inequality analysis, the parallel distributed compensation, sector of nonlinearity, and local approximation approach can reduce the complexity caused by the if-then issue of the fuzzy controller. The uncertainties in the equations are controlled using this method by breaking the fuzzy controller into smaller portions and examining the stability of each rule independently. This helps to reduce controller complexity and avoid the ``curse of dimensionality" problem that can occur with standard fuzzy controllers when the number of inputs and rules increases.

Furthermore, by approximating complex nonlinear functions with more straightforward linear or mathematical functions in different regions of the input space, the local approximation strategy can help to reduce the number of rules necessary for the fuzzy controller. Lowering the number of rules that must be evaluated can assist in simplifying the fuzzy controller even further and improve its efficiency \cite{wang2022leader}. Overall, by managing the flow of uncertainty and minimizing its complexity, these strategies can increase the robustness and efficiency of the fuzzy controller. In addition, these approaches can allow the exploration of the stability of each rule of the fuzzy controller separately. An adaptive fuzzy type II controller has been designed for the wheeled mobile robot type (2,0) \cite{ha2021adaptive}. They discuss the overall system based on the Lyapunov theory and the closed-loop stability of controllers. According to their simulation, the adaptive fuzzy type II controller performs better than the fuzzy type I controller in terms of performance. It can also reduce position error and produce smoother velocity during movements. In our approach,  the if-then rules of the fuzzy box will be mapped into parallel distributed compensation and linear matrix inequality analysis.



This paper's brief description of classic control design follows the mathematical extension of equations. Then an overview of a path by Bezier design evaluates briefly. Parallel distributed compensation, zoning, and local approximation approach are described, and then they have been added to the type I and type II Takagi-Sugeno fuzzy controllers. The controllers simulate and are tested in the MATLAB software environment in the last part. 

In brief, the contributions of this paper are summarized as follows.
\begin{itemize}
    \item Fuzzy Type I controller based on Takagai-Sugino
    \item Fuzzy Type II controller based on Takagai-Sugino
    \item Parallel Distributed Compensation
    \item Zoning and Local Approximation Approaches
\end{itemize}

The remaining of this paper is organized as follows. Section \ref{sec:related} provides a summary of the state-of-the-art studies related to our work. Section \ref{sec:problem} discusses the research problem background, and Section \ref{sec:methodology} presents our proposed approach. Next, Section \ref{sec:evaluation} details our evaluation setups and key findings. Section \ref{sec:conclusion} summarizes this work and highlights future research plans. Last, the list of acronyms used in this paper is given by Table \ref{tab:Tabel_2}. It is worth noting that T-S is the notation of the Takagi-Sugino model, and PDC stands for Parallel Distributed Compensation. We will use the term WMR instead of wheeled mobile robot type (2,0). Furthermore, it should be mentioned that the numbers used for the points, such as \ref{eq:error} or \ref{eq:points}, were intentionally picked as reasonable example values.

\begin{table}[h]
    \caption{Notations.}
    \label{tab:Tabel_2}
    \centering
    \begin{tabular}{c|c}
    \hline
        Notation & Explanation \\
        \hline
        WMR & Wheeled Mobile Robot type (2,0)\\
        LMI & Linear Matrix Inequalities \\
        PDC & Parallel Distributed Compensation \\
        T-S & Takagi-Sugino model \\
        $\upsilon$ & Linear Velocity\\
        $\omega$ & Angular Velocity \\
        Type I & Fuzzy Takagi-Sugino Type 1 Controller\\
        Type II & Fuzzy Takagi-Sugino Type 2 Controller\\
        \hline
    \end{tabular}
\end{table} \label{sec:intro}

\section{Related Work} The development of dynamic and kinematics for a wheeled mobile robot has been explored in prior work \cite{campion2011structural, parhi2011kinematic}. These have demonstrated that the set of wheeled mobile robots, despite the variety of robot designs and potential wheel configurations, can be split into five groups based on mobility and steerability \cite{parhi2011kinematic}.

In \cite{yekinni2019fuzzy}, a fuzzy logic-based type I controller for a wheeled mobile robot type (2,0) is proposed, and after designing a fuzzy logic controller, the Simulink model has been developed. However, the angle and steering are not directly controlled by the fuzzy controller with separate fuzzy rules for the left and right wheels. This issue can be solved by extending the equations and choosing angle and steering as control parameters. This modification can help reduce the navigation time employed in \cite{yekinni2019fuzzy}. They evaluate this reduction using a Mamdani controller with triangular membership functions (MFs). The result shows this strategy is more efficient than working on the left and right wheels.

A wheeled mobile robot must move smoothly and safely while tracking a path. Path simulation is only sometimes addressed \cite{yekinni2019fuzzy, ParhiPerformance, ha2021adaptive, vstefek2021optimization, xingquan2012robust}. In this work, we propose a Bezier curve to simulate different types of paths as part of our work to guarantee the various kinds of directions. Path tracking involves the geometric and timing law of wheeled mobile robots. Parhi et al. \cite{ParhiPerformance} discusses wheeled mobile robot type (2,0) mathematical equations using the kinematic model, then compares the path-tracking performance of controllers: PID and Fuzzy Logic type I. They simulated the controllers using MATLAB.  Their results show that Fuzzy Controller performs better than the PID controller, where fuzzy minimizes the Integral Square Error (ISE) during the WMR tracking.

A fuzzy logic controller with optimization is suggested in \cite{vstefek2021optimization}. Genetic algorithms (GA) are used in their architecture to ensure movement performance and reduce energy consumption for wheeled mobile robots. The earlier approaches mainly focused on energy reduction by planning the shortest path. In contrast, this approach aims to minimize the acceleration of the robot movement and energy consumption during moving from one point to the target, which can help in some applications of wheeled mobile robots. The central controller is designed based on fuzzy logic systems type I by developing a simulation in a Python environment. The fuzzy logic controller uses the experiments developed to handle the robot's navigation to a certain point. In contrast, the optimization control is generated by an auto GA, which tunes the fuzzy logic controller's fuzzy parameters like MF inputs and outputs.

In \cite{xingquan2012robust}, a tracking error base model for a wheel mobile robot, i.e., probably type (2,0), was used. A Takagi-Sugeno fuzzy controller has been considered the central controller, where varying velocity in the desired trajectory is regarded as the uncertain parameter. Then a robust controller in the form of state feedback was derived for each subsystem by solving LMI, which is the base stability analysis of our model. In the last part of the paper \cite{xingquan2012robust}, the overall controller has been constructed in the framework of parallel distributed compensation, and the Lyapunov stability of the closed-loop system has been explored. In general, the derived LMI can guarantee the stability of the controller.

Differing from the existing literature and to the best of our knowledge, this paper presents a controller based on two concepts: Fuzzy Logic and PDC. The presented controller employs the advantages of fuzzy logic and PDC to address the path tracking issue for a wheeled mobile robot of type (2,0) and improve the robot's performance in terms of stability and accuracy. While PDC controllers are notable for their robustness and potential to provide stability assurances, fuzzy logic controllers have shown to be useful in handling non-linear systems with uncertainty. The proposed controller has two stages: a fuzzy logic controller that generates control signals based on the robot's position and orientation error, and a PDC controller that optimizes these signals to improve the robot's performance. Using simulations in MATLAB, the proposed controller's performance is compared, and the results show that the type II controller is superior in path-tracking accuracy. The proposed controller can be used in various settings, including industrial automation, mobile robotics, and autonomous vehicles. \label{sec:related}

\section{Problem Abstraction and Definitions} \label{sec:problem} This section provides the definitions and research problem abstraction for this paper's study of a wheeled mobile robot (WMR). It is considered that the WMR under study has a rigid frame and horizontally moving wheels that do not deform. The local ($X1, X2$) and reference ($I1, I2$) coordinate systems depict the robot's position on the surface. A three-dimensional vector is used to indicate the location of the WMR in the coordinate systems. The study assumes that there are no obstacles, and therefore, point-to-point control is used. Yet, even in an environment containing obstacles, any path can be generated using Bezier curves. The third wheel, the caster, keeps the WMR balanced while two motors drive the two fixed wheels. 

The output variables are tied to $x$, $y$, and the control inputs are linear and rotational velocities. The paper provides equations for the wall or path tracking challenge, where the WMR's objective is to follow the wall or path at a specific speed. A third-order Bezier curve with four control points is employed in this study. The Bezier curve is used to construct various pathways. The first and last control points are fixed, and the path's shape is modified by adjusting the second and third control points.

\begin{figure}[h]
    \centering
    \includegraphics[width=7cm]{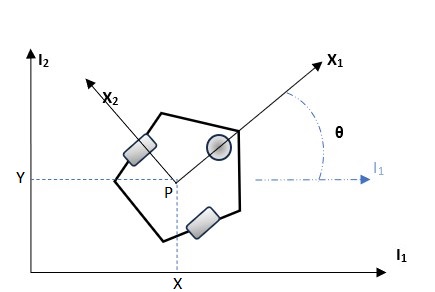}
    \caption{WMR position in local and reference frame.}
    \label{fig:position}
\end{figure}

\subsection{Wheeled Mobile Robot}
In this paper, we will ignore the wheel slippage through displacement. Furthermore, we assume that the WMR studied in this paper, Fig. \ref{fig:Type_2_0}, comprises a rigid frame (body) with non-deformable wheels that move horizontally. 

The robot's position on the surface is defined in Fig. \ref{fig:position}, which includes two coordinate systems, the Local $\{X_1, X_2\}$, which is connected to the robot's firm and the Reference $\{I_1, I_2\}$ with index L and R respectively. Consider the location of WMR like a point P in the reference coordinate system, where the point $P$ is the geometric center of gravity of the WMR; a three-dimension vector can depict the WMR's position in the reference coordinate system:
\begin{equation}
P_R=[X,Y,\theta]^T
\end{equation}
The $\theta$ is the orientation of the WMR; the angle that travels in the direction of the vertex of triangular from the $X_1-$axis to the $I_1-$axis in the reference coordinate system. It is necessary to interpret any action of the WMR in both coordinate systems.

The orthogonal rotation matrix R($\theta$) maps the displacement in the direction of the reference coordinate system to the displacement in the local coordinate system:
\begin{equation}
    P_R=R(\theta) P_L
\end{equation}
where rotation matrix is:
\begin{equation}
    R(\theta)=\begin{pmatrix}
        \cos\theta & \sin\theta & 0 \\ 
        -\sin\theta & \cos\theta & 0 \\
        0 & 0 & 1
    \end{pmatrix}
\end{equation}

\begin{figure}[h]
    \centering
    \includegraphics[width=8cm]{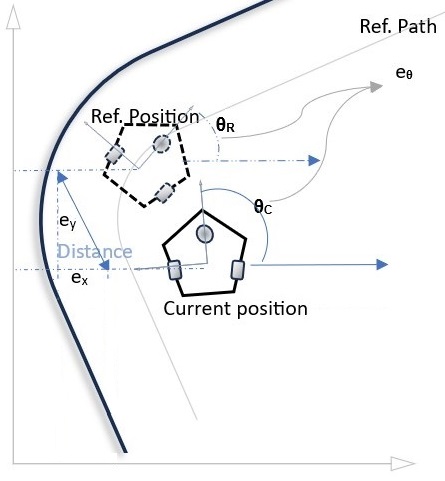}
    \caption{Moving of WMR and variables.}
    \label{fig:moving}
\end{figure}

Our design is based on an obstacle-free environment that needs point-to-point control. However, the representation of paths by Bezier curves allows us to generate any path, even if displacement through an environment with obstacles is needed. The required path can be designed using a Bezier curve after a fast scan of the surroundings. Two motors control the two fixed wheels, and the third wheel, known as the caster, maintains the balance of WMR. If the indexes $r$ and $l$ show right and left wheel properties, respectively, $L_W$ is half of WMR's width and $R_W$ the radius of wheels. Therefore:
\begin{equation}
    \begin{cases}
    \upsilon_r=\omega_r.R_W \\
    \upsilon_l=\omega_l.R_W\\
    \upsilon=\frac{1}{2}(\upsilon_l+\upsilon_r)\\
    \omega=\frac{1}{2L_{W}}(\upsilon_r-\upsilon_l)
    \end{cases}
\end{equation}
where $\upsilon$ and $\omega$  are the linear and rotational velocities. Fig. \ref{fig:moving} shows the WMR in both current and reference positions. The kinematic model of the displacement on the coordinate system can be expressed as follows:
\begin{equation}
    \begin{cases}
    \dot x=\upsilon\cos\theta \\ \dot y=\upsilon\sin\theta \\ \dot \theta =\omega
    \end{cases}
\end{equation}
The control inputs are the linear and rotational velocities, and the output variables are connected to $x, y$, and $\theta$. 
The actual position \(P_C=\begin{bmatrix}  x_C & y_C & \theta_C \end{bmatrix}^T\) and the target position \(P_R=\begin{bmatrix}  x_R & y_R & \theta_R \end{bmatrix}^T\) are used as the current and reference positions of the WMR in our design. The WMR’s goal is to follow the wall or path at a certain speed, so the following equations appear for the wall or path tracking task.
\begin{equation}
    \begin{cases}
    \dot x_R=\upsilon_R\cos\theta_R, \\ \dot y_R=\upsilon_R\sin\theta, \\ \dot \theta_R =\omega_R.
    \end{cases}
\end{equation}
As shown in Fig. \ref{fig:moving}, the definition of position error \(e_P=\begin{bmatrix}e_x &  e_y & e_\theta\end{bmatrix}^T\) using the previous definitions, gives the following equation:
\begin{equation}
    \begin{bmatrix}e_x \\  e_y \\ e_\theta\end{bmatrix}
    =\begin{bmatrix}
        \cos\theta & \sin\theta & 0 \\ 
        -\sin\theta & \cos\theta & 0 \\
        0 & 0 & 1
    \end{bmatrix} [P_R-P_C]
\end{equation}
The rotation matrix sends the position error to the local coordinates. Consequently, when $P_C$ equals $P_R$, the errors are zero, which means the WMR performs its task perfectly.
\subsection{Path Modeling}

\begin{figure*}[h]
    \centering
    \includegraphics[width=\textwidth]{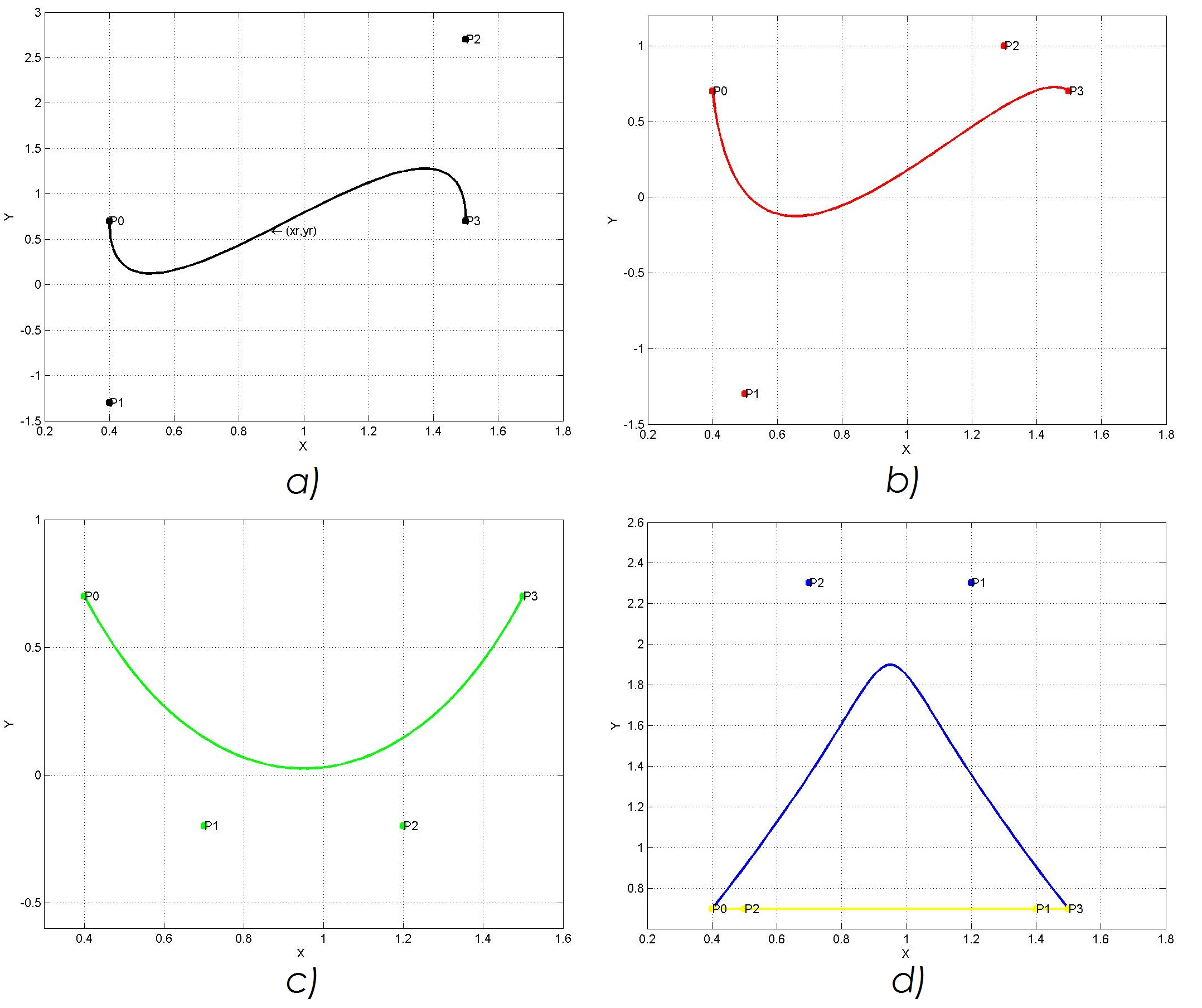}
    \caption{Bezier curve; Changing the control points.}
    \label{fig:Paths}
\end{figure*}

The Bezier curve is a type of parametric curve that is used in graphics and other fields. In this research, we use a third-order Bezier curve to design different paths. In future research, we can test and run the controller for WMR with other functions such as path following, player robot, wall tracking, etc. A 3rd order Bezier curve with four control points is seen in Fig. \ref{fig:Paths}. Fig. \ref{fig:Paths} (a to d) depicts how several pathways can be reached by changing only two control points on a third-order Bézier curve. The initial and last control points are fixed, and the path shape changes by changing the second and third control points. Now, a WMR can follow the paths depicted in Fig. \ref{fig:Paths}. The path (in Fig. \ref{fig:Paths}.a) starts from the initial point $(x_i,y_i)=(0.4,0.7)$ with $\theta_i=-90$ and ends at the final position $(x_f,y_f)=(1.5,0.7)$ with $\theta_f=-90$ arrives. It should be considered that the maximum available rotational velocity is in the rotation points.

Moreover, the WMR can accelerate or decelerate before these points of rotation. The following forms for the linear velocity and rotational velocity profiles are accessible when the main velocity relation for the speed limit is utilized. After the initial time $\Delta t$, the WMR moves to the neighborhood of the second control point, and $\Delta s$ is the displacement length. The $\upsilon(\eta)$ notes average velocity during this path, therefore in the case of infinite $\Delta s$, WMR's displacement is expressed as follows:
\[\Delta s(\eta)=\upsilon(\eta)\Delta t\]

Here, the average velocity is:
\[\upsilon(\eta)=\frac{1}{\Delta t}\sqrt{\Delta x(\eta)^2+\Delta y(\eta)^2}\]
And when $\Delta t \rightarrow 0$, the average velocity is close to the instantaneous speed. We arrive at the equation below, which is the length of the curvature of the path.
\[ds(\eta)=\sqrt{x\prime_R(\eta)^2+y\prime_R(\eta)^2} d(\eta)\]
Where $x\prime_R(\eta)$ and $y\prime_R(\eta)$ are derivatives $x$ and $y$ respect to $\eta$. As can be seen in the following figures, the length of the path is obtained using an integral square numerical pattern from the equation:
\[t=\int_{0}^{1}\frac{\sqrt{x\prime_R(\eta)^2+y\prime_R(\eta)^2}}{\upsilon_r(\eta)}\,d\eta \]
In this research, we assume velocity is unlimited; therefore, the displacement time is always 1 second.

\begin{figure*}[h]
    \centering
    \includegraphics[width=\textwidth]{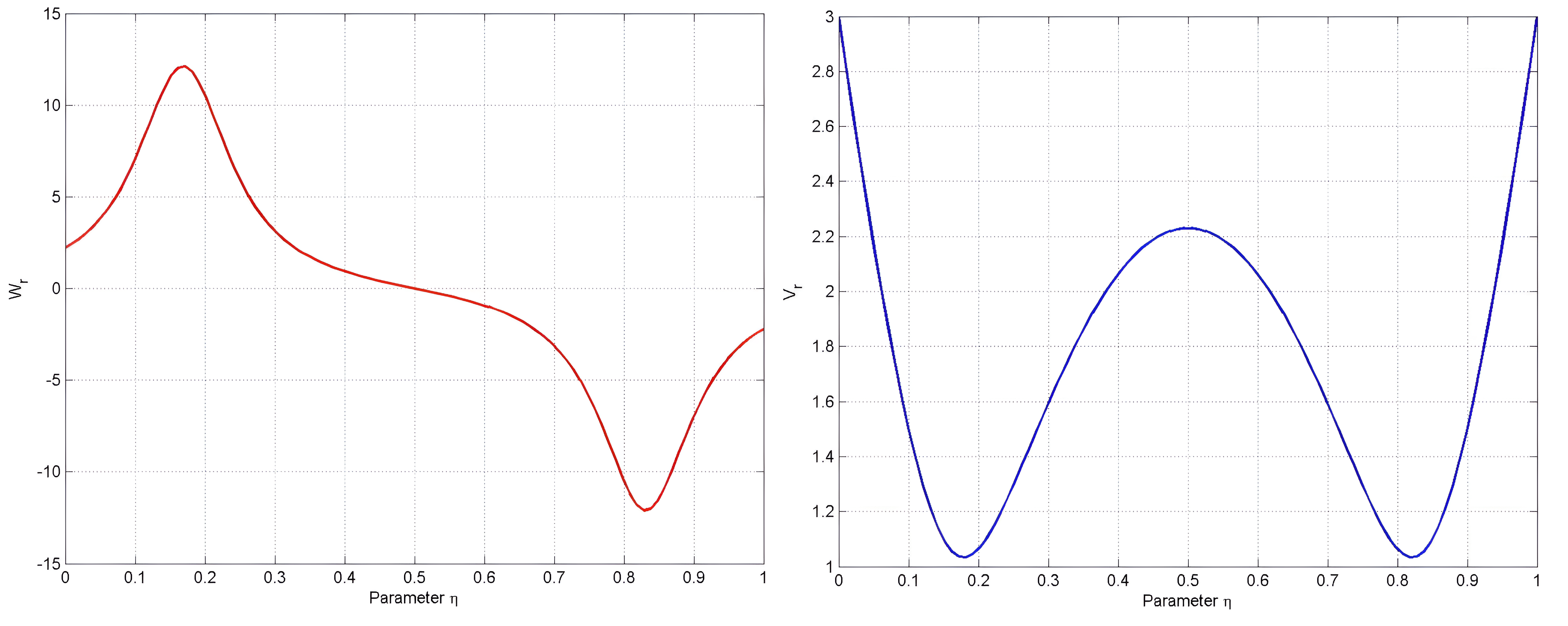}
    \caption{Rotational and linear velocity profiles (for 6. a).}
    \label{fig:profiles}
\end{figure*}


According to the profile shape of linear and rotational velocities of the path (Fig. \ref{fig:profiles}), the minimum and maximum reference linear velocities are 1.0334 and 3 $\frac{m}{s}$, respectively. Moreover, the minimum and maximum reference rotational velocities are 12.0968 and 12.0968 $\frac{rad}{s}$, respectively.



\section{Methodology} \label{sec:methodology} The Wheeled Mobile Robot (WMR) is propelled toward the corresponding reference positions by the controller outputs, which are velocities.
The controller's goal is to reduce the robot's position error to zero, which also is the system’s operating point. Moreover, the system inputs are the reference position $P_R$ and the reference velocities $Q_R=(\upsilon_R,\omega_R)$, which are time functions. With the previous equations and considering these topics, we can provide specific equations for every function of WMR. As an example, suppose a WMR must follow a wall by maintaining the distance $Y_0$ (or $X_0$); therefore, the $e_x$ (or $e_y$) always is zero (in most studies, this assumption has been made - ref). By changing the variable $y=Y-Y_0$ (or $x=X-X_0$), the same control objectives return, and the same working points appear.

\begin{figure}[h]
    \centering
    \includegraphics[width=8cm]{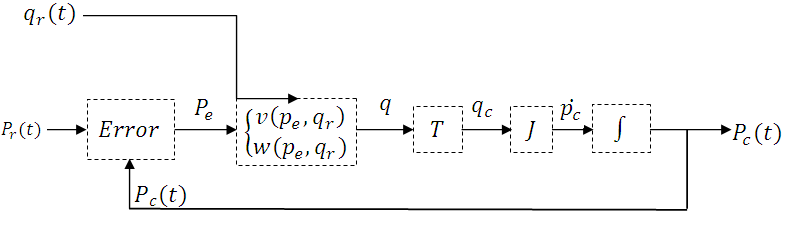}
    \caption{WMR architecture.}
    \label{fig:architecture}
\end{figure}

\begin{figure*}[h]
    \centering
    \includegraphics[width=\textwidth]{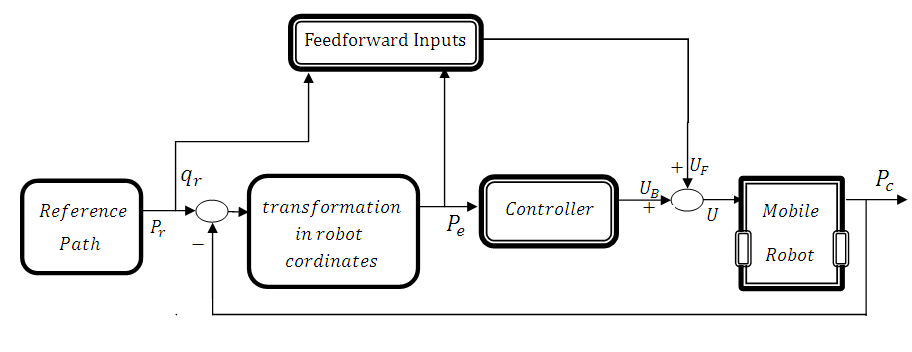}
    \caption{Controller architecture, including both feedforward and feedback components.}
    \label{fig:controller}
\end{figure*}

The proposed overall WMR architecture is shown in Fig. \ref{fig:architecture}. The first component calculates the error between the current and reference positions. The controller runs in the second component, which calculates the target velocities using position errors and reference velocities. The third component, T, is the WMR hardware, which converts target speeds to actual WMR speeds. This conversation is related to robot dynamics and is outside the scope of this research; we will assume $Q_C=(\upsilon_c,\omega_c)=(\upsilon,\omega)=Q$. The fourth component is the kinematic matrix mentioned in previous equations, which provides the derivative of the current position. The last component integrates this to generate position. The second component is the main focus of this research. The position error model can be written as follows:

\begin{equation}
    \dot e_P=
    \begin{bmatrix}\dot e_x \\  \dot e_y \\ \dot e_\theta\end{bmatrix}=
    \begin{bmatrix} e_y \omega _c-\upsilon_c + \upsilon_R \cos e_\theta\\ -e_x \omega _c + \upsilon_R \cos e_\theta\\ \omega _R - \omega _c \end{bmatrix}
\end{equation}

where $\upsilon_R$ is the reference linear velocity and $\omega _R$ is the reference rotational velocity. Rewriting the above equation results in the following:
\begin{equation}
    \begin{bmatrix} \dot e_x \\  \dot e_y \\ \dot e_\theta\end{bmatrix}=
    \begin{bmatrix} \cos e_\theta & 0\\ \sin e_\theta & 0\\ 0 & 1 \end{bmatrix}
    \begin{bmatrix} \upsilon_R\\ \omega_R \end{bmatrix}-
    \begin{bmatrix} -1 & e_x\\  0 & -e_y\\ 0 & -1 \end{bmatrix}
    \begin{bmatrix} \upsilon_c\\ \omega _c \end{bmatrix}
\end{equation}
Fig. \ref{fig:controller} shows the architecture of our controller, including both feedforward and feedback components as the following equations show.
\begin{equation}
        \begin{bmatrix} \upsilon_c \\ \omega _c \end{bmatrix}=U_F+U_B
\end{equation}
where:
\begin{equation}
    U_F=\begin{bmatrix} \upsilon_R \cos e_\theta & \omega_R \end{bmatrix}^T
\end{equation}
$U_F$ is the feedforward control vector and $U_B$ is the feedback controller as explained
in the following sections. The control equation can be expressed in the following nonlinear form.
\begin{equation}
    \begin{bmatrix}\dot e_x \\  \dot e_y \\ \dot e_\theta\end{bmatrix}=
    \begin{bmatrix} 0 & \omega_R & 0\\ -\omega_R & 0 & \upsilon_r \cos e_\theta\\ 0 & 0 & 1 \end{bmatrix}
    \begin{bmatrix} e_x \\ e_y \\ e_\theta\end{bmatrix}-
    \begin{bmatrix} -1 & e_x\\  0 & -e_y\\ 0 & -1 \end{bmatrix} U_B
\end{equation}



\subsection{Controller}
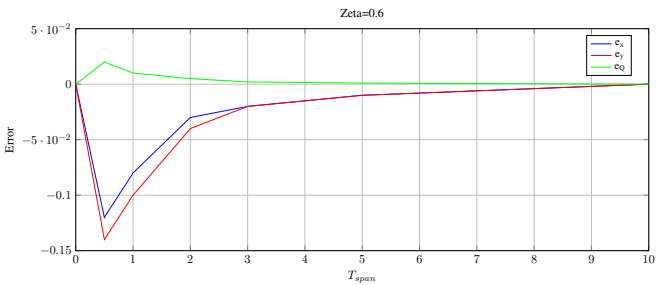
\begin{figure}[h]
    \centering
    \resizebox{\columnwidth}{!}{
    \begin{tikzpicture}
        \begin{axis}[
            width=\textwidth,
            height=8cm,
            xlabel={$T_{span}$},
            ylabel={Error},
            title={Zeta=0.6},
            legend pos=north east,
            legend style={font=\small},
            grid=both,
            ymin=-0.15, ymax=0.05,
            xmin=0, xmax=10
        ]
            \addplot[blue, thick] coordinates {
                (0,0)
                (0.5,-0.12)
                (1,-0.08)
                (2,-0.03)
                (3,-0.02)
                (5,-0.01)
                (10,0)
            };
            \addlegendentry{e\textsubscript{x}}
            
            \addplot[red, thick] coordinates {
                (0,0)
                (0.5,-0.14)
                (1,-0.1)
                (2,-0.04)
                (3,-0.02)
                (5,-0.01)
                (10,0)
            };
            \addlegendentry{e\textsubscript{y}}
            
            \addplot[green, thick] coordinates {
                (0,0)
                (0.5,0.02)
                (1,0.01)
                (2,0.005)
                (3,0.002)
                (5,0.001)
                (10,0)
            };
            \addlegendentry{e\textsubscript{Q}}
        \end{axis}
    \end{tikzpicture}
    }
    \caption{Classic controller for 6. a.}
    \label{fig:classic}
\end{figure}

The classical controller starts with linearization around the point $e_x=e_y=e_\theta$ and $U_{B1}=U_{B2}$. The model calculates the following:
\[\dot e= A e + B U_B\] 
that
\[ A=\begin{bmatrix} 0 & \omega_R & 0\\ -\omega_R & 0 & \upsilon_R \\ 0 & 0 & 0 \end{bmatrix}\]
\[B= \begin{bmatrix} -1 & 0 \\0 & 0 \\ 0 & -1 \end{bmatrix}\]
And the controllability matrix is:
\begin{equation}
   \begin{bmatrix} A \\ AB \\ A^2B \end{bmatrix}^T = 
   \begin{bmatrix} 1 & 0 & 0 & 0 & - \omega_R^2 & \epsilon_R\omega_R \\
    0 & 0 & 0 & - \omega_R^2 & \epsilon_R\omega_R  & 0 &\\
    0 & 1 & 0 & 0 & 0 & 0
    \end{bmatrix}
\end{equation}
Therefore, with non-zero $\upsilon_R$ and $\omega_R$, a constant matrix shows a straight-line
route, and with a complete rank matrix, the route is part of a circle. Consider a feedback controller as $\upsilon =k.e$ :
\begin{equation}
   \begin{bmatrix} \upsilon \\ \omega \end{bmatrix} = 
   \begin{bmatrix} k_{11} & k_{12} & k_{13} \\
    k_{21} & k_{22} & k_{23} \\
    \end{bmatrix}
    \begin{bmatrix} e_x \\ e_y \\ e_\theta\ \end{bmatrix}
\end{equation}
where $k_{11}=k_1$, $k_{22}=-sign(\upsilon_R)k_2$ and $k_{23}=-k_3$ the poles of the $det(SI-A+Bk)$ are $(s+2\zeta\omega_n)(s^2+2\zeta\omega s+\omega_n^2)$ therefore:
\begin{equation}
    \begin{cases}
    k_1=k_3=2\zeta\omega_n \\
    k_2=\frac{\omega_n^2-\omega_R^2}{|\upsilon_R|}\\
    \end{cases}
\end{equation}
To prevent the second coefficient from becoming infinity, when $\upsilon_r$ goes toward zero, we define $k_2=\upsilon_R.g$, which leads $k_1=k_2=k_3=2\sqrt{\omega_R^2+g.\upsilon_R^2}$. Therefore, with $\zeta=0.6$ and $g=0.1$ for a simple classical controller, the results are as shown in Fig. \ref{fig:classic}.

\subsection{Approaches: Sector of Nonlinearity, local approximation, and PDC} 
Experiences and knowledge of human expertise are often used to develop fuzzy rules and membership functions. In another approach, databases that include previously collected data on the system's behavior or real-world scenarios can be used to extract information for the fuzzy controller's design. However, using mathematical methods rather than human expertise or databases allows for a more objective and scientific approach to constructing the fuzzy controller. This is the approach adopted here. The first step is the fuzzification of inputs.

The idea of the sector of nonlinearity in fuzzy model structures was first mentioned in \cite{nguyen2019fuzzy}. Consider a simple nonlinear example $\dot x=f(x(t))$ such that $ f(0)=0$. The main goal is to find a general area that: 
\begin{equation}
\dot x =f(x(t))\in [a_1,a_2]x(t)
\end{equation}
This approach ensures the exact structure of the fuzzy model. But sometimes, it isn’t easy to find general areas for nonlinear systems which satisfy the above condition. In this case, the area can be considered locally.

The second approach is called local approximation in fuzzy space. The local approximation approach reduces the number of rules. The basis of this approach is to approximate nonlinear terms by selecting linear terms correctly.  It is important to note that the number of rules is directly related to the complexity of the analysis and design. However, designing control rules based on the local approximation may not guarantee the stability of the main nonlinear system under the control rules. One way to solve this problem is to introduce a robust control design.

Parallel compensation provides a procedure for designing a fuzzy controller from a fuzzy T-S model \cite{wang2021survey}. In parallel compensation design, each control rule is designed from the corresponding rule in the fuzzy T-S model. Moreover, the fuzzy controller shares the same fuzzy sets of fuzzy models in the initial section. The structure of the T-S controller type I distributed compensation
method is as follows:
\\Rule $i^th$  of the controller:\\
if $z_1 (t)$ is  $M_{i1}$ and $\cdots$ and $z_p (t)$ is $M_{ip}$  then
\begin{equation} \label{eq:PDC1}
u_B(t)=-F_i x(t)
\end{equation}
\begin{equation} \label{eq:PDC2}
u_B(t)=-\frac{\sum_{i=1}^{r} W_i(Z(t))F_ix(t)}{\sum_{i=1}^{r}W_i(Z(t))}=-\sum_{i=1}^{r} h_i(Z(t))F_ix(t)
\end{equation}
The design of the parallel distributed compensation controller determines the local feedback weights of  $F_i$ in consecutive sections.

\begin{figure*}[h]
    \centering

    \begin{subfigure}[b]{0.48\textwidth}
        \centering
        \begin{tikzpicture}
            \begin{axis}[
                width=\textwidth,
                height=6cm,
                xlabel={$W_r$},
                ylabel={},
                title={Membership Functions E1 (Blue Line) and E2 (Red Line)},
                legend pos=north east,
                legend style={font=\small},
                grid=both,
                ymin=0, ymax=1,
                xmin=-15, xmax=15
            ]
                \addplot[blue, thick] coordinates {
                    (-15,1)
                    (15,0)
                };
                \addlegendentry{E1}
                
                \addplot[red, thick, dashed] coordinates {
                    (-15,0)
                    (15,1)
                };
                \addlegendentry{E2}
            \end{axis}
        \end{tikzpicture}
    \end{subfigure}
    \hfill
    \begin{subfigure}[b]{0.48\textwidth}
        \centering
        \begin{tikzpicture}
            \begin{axis}[
                width=\textwidth,
                height=6cm,
                xlabel={$e_y$},
                ylabel={},
                title={Membership Functions P1 (Red Line) and P2 (Blue Line)},
                legend pos=north east,
                legend style={font=\small},
                grid=both,
                ymin=0, ymax=1,
                xmin=-0.2, xmax=0.2
            ]
                \addplot[red, thick, dashed] coordinates {
                    (-0.2,1)
                    (0.2,0)
                };
                \addlegendentry{P1}
                
                \addplot[blue, thick] coordinates {
                    (-0.2,0)
                    (0.2,1)
                };
                \addlegendentry{P2}
            \end{axis}
        \end{tikzpicture}
    \end{subfigure}

    \vspace{1em}

    \begin{subfigure}[b]{0.48\textwidth}
        \centering
        \begin{tikzpicture}
            \begin{axis}[
                width=\textwidth,
                height=6cm,
                xlabel={$V_r \sin(e_x)$},
                ylabel={},
                title={Membership Functions F1 (Blue Line) and F2 (Red Line)},
                legend pos=north east,
                legend style={font=\small},
                grid=both,
                ymin=0, ymax=1,
                xmin=2, xmax=6
            ]
                \addplot[blue, thick] coordinates {
                    (2,1)
                    (6,0)
                };
                \addlegendentry{F1}
                
                \addplot[red, thick, dashed] coordinates {
                    (2,0)
                    (6,1)
                };
                \addlegendentry{F2}
            \end{axis}
        \end{tikzpicture}
    \end{subfigure}
    \hfill
    \begin{subfigure}[b]{0.48\textwidth}
        \centering
        \begin{tikzpicture}
            \begin{axis}[
                width=\textwidth,
                height=6cm,
                xlabel={$e_x$},
                ylabel={},
                title={Membership Functions G1 (Red Line) and G2 (Blue Line)},
                legend pos=north east,
                legend style={font=\small},
                grid=both,
                ymin=0, ymax=1,
                xmin=-0.2, xmax=0.2
            ]
                \addplot[red, thick, dashed] coordinates {
                    (-0.2,1)
                    (0.2,0)
                };
                \addlegendentry{G1}
                
                \addplot[blue, thick] coordinates {
                    (-0.2,0)
                    (0.2,1)
                };
                \addlegendentry{G2}
            \end{axis}
        \end{tikzpicture}
    \end{subfigure}

    \caption{The membership functions of Type I.}
    \label{fig:membership_1}
\end{figure*}
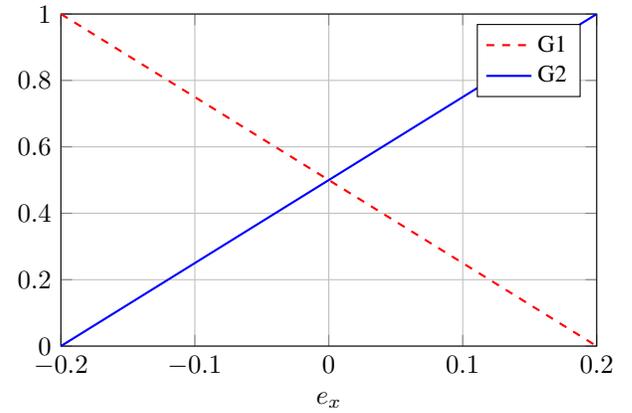

\subsection{Type I}
Our design is based on the T-S model and Type I and Type II fuzzy logic. The Mamdani, Sugeno, and T-S fuzzy controllers are the three most well-known fuzzy models, and each type of fuzzy logic can be classified as either type I or II. Due to the abundance of readily accessible resources, we will not go into detail. 
As typical in a feedback controller, the process output is compared to a reference. Based on the existing differences and control strategy, the controller applies the required signal to the process.

\underline{Takagi-Sugeno:}
The $i^{th}$ rule of fuzzy T-S models for our controller is:
\\Rule $i^th$ of the ``IF-THEN" rules set:
\\if $z_1 (t)$ is  $M_{i1}$ and $\cdots$ and $z_p (t)$ is $M_{ip}$  then

\begin{align}
    \begin{cases}
        \dot x=A_ix(t)+B_iu(t)\\
        y(t)=C_ix(t)
    \end{cases}
    ~~i=1, \cdots,r
\end{align}
Where $M_{ip}$ are fuzzy sets, and $r$ is the number of rules; $x(t)\in R^n$ is the state vector, $u(t) \in R^m $ is the input vector, and $y(t) \in R^q $ is the output vector; also $A_i \in R^{n.n}$, $B_i \in R^{n.m} $, $C_i \in R^{n.n}$, and \(Z_1(t), Z_2(t),\cdots, ~and~ Z_p(t)\) are linguistic variables that can be functions of state variables, external disturbances, or time. We use $Z(t)$ to represent the vector containing all the distinct elements $Z_1 (t),Z_2 (t),\cdots, $ and $ Z_p(t)$. Each linear conceptual equation represented by $A_i x(t)+B_i u(t)$ is called a subsystem, and the equations of the fuzzy system are as follows:
\begin{equation}
\dot x(t)=\sum_{i=1}^{r} h_i(Z(t)){A_ix(t)+B_iu(t)}
\end{equation}
\begin{equation}
\dot y(t)=\sum_{i=1}^{r} h_i(Z(t))C_ix(t)
\end{equation}
where
\begin{equation}
W_i(Z(t))=\prod_{j=1}^{p}M_{ij}z_j(t)
\end{equation}
\begin{equation}
h_i(Z(t))=\frac{W_i(Z(t))}{\sum_{i=1}^{r}W_i(Z(t))}
\end{equation}
the component $M_{ij} (z_j (t)) $ is the degree of membership of $z_j (t)$ in $M_{ij} $. Since:
\begin{equation}
    \begin{cases}
       \sum_{i=1}^{r}W_i(Z(t))>0 \\ 
       W_i(Z(t))\geq0
    \end{cases}
\end{equation}
For all times, we have:
\begin{equation}
    \begin{cases}
       \sum_{i=1}^{r}h_i(Z(t))=1 \\ 
       h_i(Z(t))\geq 0
    \end{cases}
\end{equation}
To achieve the fuzzy model from the nonlinear model of the system, two ideas of the sector of nonlinearity, local approximation, or a combination of both can help \cite{wang1995analytical}.
Since the system state-equation are:
\begin{equation}\label{eq:main}
    \begin{bmatrix}\dot e_x \\  \dot e_y \\ \dot e_\theta\end{bmatrix}=
    \begin{bmatrix} 0 & \omega_R & 0\\ \omega_R & 0 & \upsilon_r \sin e_\theta\\ 0 & 0 & 1 \end{bmatrix}
    \begin{bmatrix} e_x \\ e_y \\ e_\theta\end{bmatrix}-
    \begin{bmatrix} 1 & -e_x\\  0 & e_y\\ 0 & 1 \end{bmatrix} U_B
\end{equation}
Using the T-S approach, the nonlinear state-space is defined with the T-S model based on the IF-THEN rules, i.e., in the rule section, each rule shows a linear state-space model. For variables, the following range of errors has been considered: 
\begin{equation}\label{eq:error}
    \begin{cases}
        |e_x| \leq 0.2 \\
        |e_y| \leq 0.2 \\
        |e_\theta| \leq \frac{\pi}{2} 
    \end{cases}
\end{equation}
Where the linguistic variables are:
\begin{equation}
    \begin{matrix}\label{eq:zoning}
        z_1 = \omega_R,  &
        z_2 = \upsilon_R \sin e_\theta &
        z_3 = e_y, &
        z_4 = e_x
    \end{matrix}
\end{equation}
for $z_1 (t)$, the zoning is:
\begin{equation} \label{eq:minmax}
    \begin{cases}
        e_1 =max(z_1(t)) \\
        e_2 = min(z_1(t))
    \end{cases}
\end{equation}
And in the defined range:
\begin{equation}
z_1(t)=\sum_{i=1}^{2} E_i(z_i(t))e_i
\end{equation}
Also, regarding memberships functions in Fig. \ref{fig:membership_1}, we have the following:
\begin{equation}
E_1(z_1(t))+E_2(z_1(t))=1
\end{equation}
Therefore:
\begin{equation}
    \begin{cases}
        E_1(z_1(t)) =\frac{z_1(t)-e_2}{e_1-e_2} \\
        E_2(z_1(t))= \frac{e_1-z_1(t)}{e_1-e_2}
    \end{cases}
\end{equation}
This means there are two membership functions for the first variable. The designer chooses the number of these functions, but increasing the number of membership functions may lead to not finding a stable controller. For other variables, zoning operates in the same way. In Fig. \ref{fig:membership_1}, we project membership functions for $z_1, z_2, z_3$, and $z_4 $.
\begin{equation}
    \begin{cases}
        F_1(z_1(t)) =\frac{z_2(t)-f_2}{f_1-f_2} \\
        F_2(z_1(t))= \frac{e_1-z_2(t)}{f_1-f_2}
    \end{cases}
\end{equation}
\begin{equation}
    \begin{cases}
        G_1(z_1(t)) =\frac{z_3(t)-e_2}{g_1-g_2} \\
        G_2(z_1(t))= \frac{g_1-z_3(t)}{g_1-g_2}
    \end{cases}
\end{equation}
\begin{equation}
    \begin{cases}
        P_1(z_1(t)) =\frac{z_4(t)-p_2}{p_1-p_2} \\
        P_2(z_1(t))= \frac{p_1-z_4(t)}{p_1-p_2}
    \end{cases}
\end{equation}

Fig. \ref{fig:membership_1} shows the membership functions for the linguistic variables.


IF-THEN rule box makes special channels for every input and lets the controller pick multiple channels. Therefore, for our model, the result of each rule is equal to the state space of the system with a degree of effect that the rule has. This decision-making is the central part of the controller and plays a vital role in its stability. In other words, with such controllers, instead of one stable point in each condition, some stable points are mixed with a degree of effectiveness so that the displacement can be smoother and prevent sharp movement. Now, considering the fuzzy membership functions and the linguistic variables, the IF-THEN rules are: 
\\ Rule 1: IF $ \begin{cases}
     z_1 \quad is \quad E_1 \\ z_2 \quad is \quad F_1 \\ z_3 \quad is  \quad G_1 \\ z_4 \quad is \quad P_1 \end{cases} $  THEN: 
\begin{equation}
    \begin{bmatrix}\dot e_x \\  \dot e_y \\ \dot e_\theta\end{bmatrix}=
    \begin{bmatrix} 0 & e_1 & 0\\ -e_1 & 0 & f_1\\ 0 & 0 & 1 \end{bmatrix}
    \begin{bmatrix} e_x \\ e_y \\ e_\theta\end{bmatrix}+
    \begin{bmatrix} -1 & g_1\\  0 & -p_1\\ 0 & -1 \end{bmatrix} U_B
\end{equation}
Rule 16: IF $\begin{cases}z_1 \quad  is \quad E_2 \\ z_2  \quad is \quad F_2 \\ z_3 \quad is \quad G_2 \\ z_4 \quad is \quad P_2 \end{cases} $ THEN: 
\begin{equation}
    \begin{bmatrix}\dot e_x \\  \dot e_y \\ \dot e_\theta\end{bmatrix}=
    \begin{bmatrix} 0 & e_2 & 0\\ -e_2 & 0 & f_2\\ 0 & 0 & 1 \end{bmatrix}
    \begin{bmatrix} e_x \\ e_y \\ e_\theta\end{bmatrix}+
    \begin{bmatrix} -1 & g_2\\  0 & -p_2\\ 0 & -1 \end{bmatrix} U_B
\end{equation}
Our rule box has 16 rules which means there is no eliminated rule. Moreover, the output based on the previous equation is:
\begin{equation}
\sum_{i=1}^{2} \sum_{j=1}^{2} \sum_{k=1}^{2} \sum_{m=1}^{2} E_iz_1(t)F_jz_2(t)G_kz_3(t)P_lz_4(t)
\end{equation}
The PDC matches each rule of T-S to the corresponding rule, i.e., the number of rules remains the same. In addition, the firing conditions of rules are identical to the T-S model. The PDC equations \ref{eq:PDC1}, \ref{eq:PDC2} and the Lyapunov stability \cite{klanvcar2007tracking, wang1995analytical} our controller can achieve by solving the LMI problem given the condition of finding the values of $X>0$ and $M_i$ satisfies the following relations:
\[-XA_i^T-A_i X+M_i^T B_i^T+B_i M_i>0\]
\[-XA_i^T-A_i X-X A_j^T-A_j X + M_j^T B_i^T+\]
\[B_i M_j+M_i^T B_j^T+B_j M_i \geq 0\]
for \begin{equation}
i<j \quad  h_i \cap h_j \neq \varnothing
\end{equation}
While \(M_i=F_iX\) and \(X=P^{-1}\)
i.e., we need to find a positive defined matrix like P that satisfies the Lyapunov conditions.

\begin{figure*}[h]
    \centering
    \includegraphics[width=\textwidth]{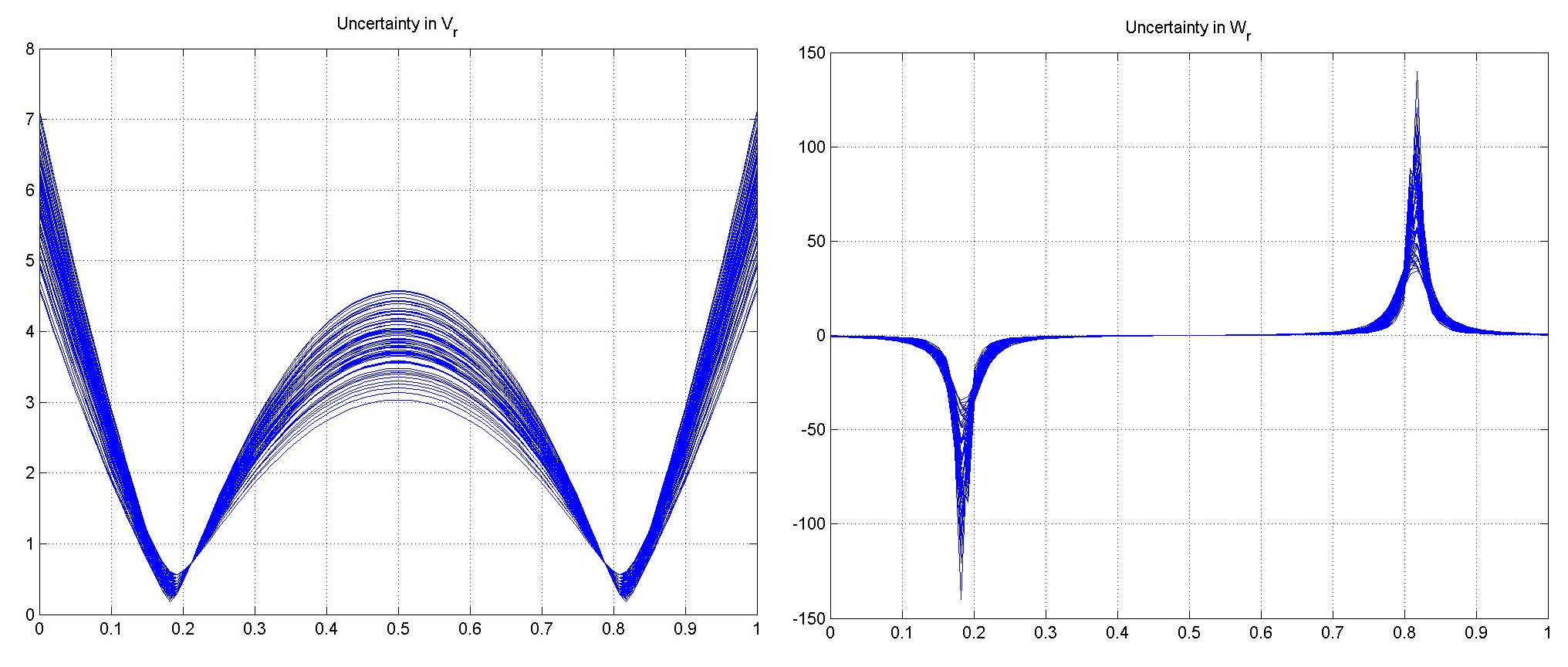}
    \caption{Uncertainty profiles for changes in $\theta_i$ and $\theta_f$.}
    \label{fig: uncertainty_theta}
\end{figure*}

\begin{figure*}[h]
    \centering
    \includegraphics[width=\textwidth]{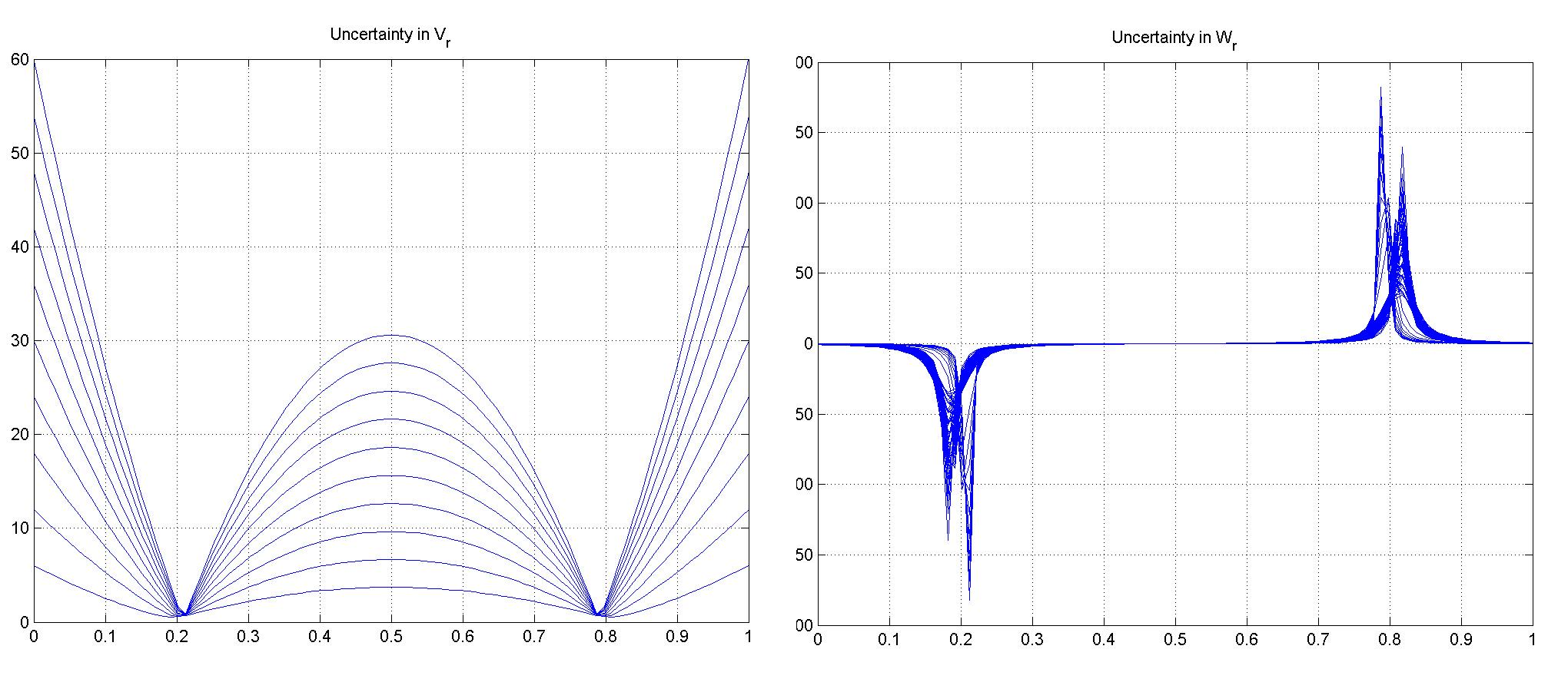}
    \caption{Uncertainty profiles for changes in $d_i$ and $d_f$.}
    \label{fig: uncertainty_d}
\end{figure*}

\subsection{Type II}
As we mentioned, by changing the parameters of the control points, the linear and angular velocities have much uncertainty, so we will investigate the effect of these factors as a primary source of uncertainty type II. Changing the angle $\theta_i$ and $\theta_f$ from the $\pi/4$ to $\pi/3$, leads the velocity profiles to have uncertainty as shown in Fig. \ref{fig: uncertainty_theta}, and for $d_i $ and $d_f$ changes from 1 to 10; velocity profiles have uncertainty as shown in Fig. \ref{fig: uncertainty_d}.

\begin{figure*}[h]
    \centering
    \includegraphics[width=\textwidth]{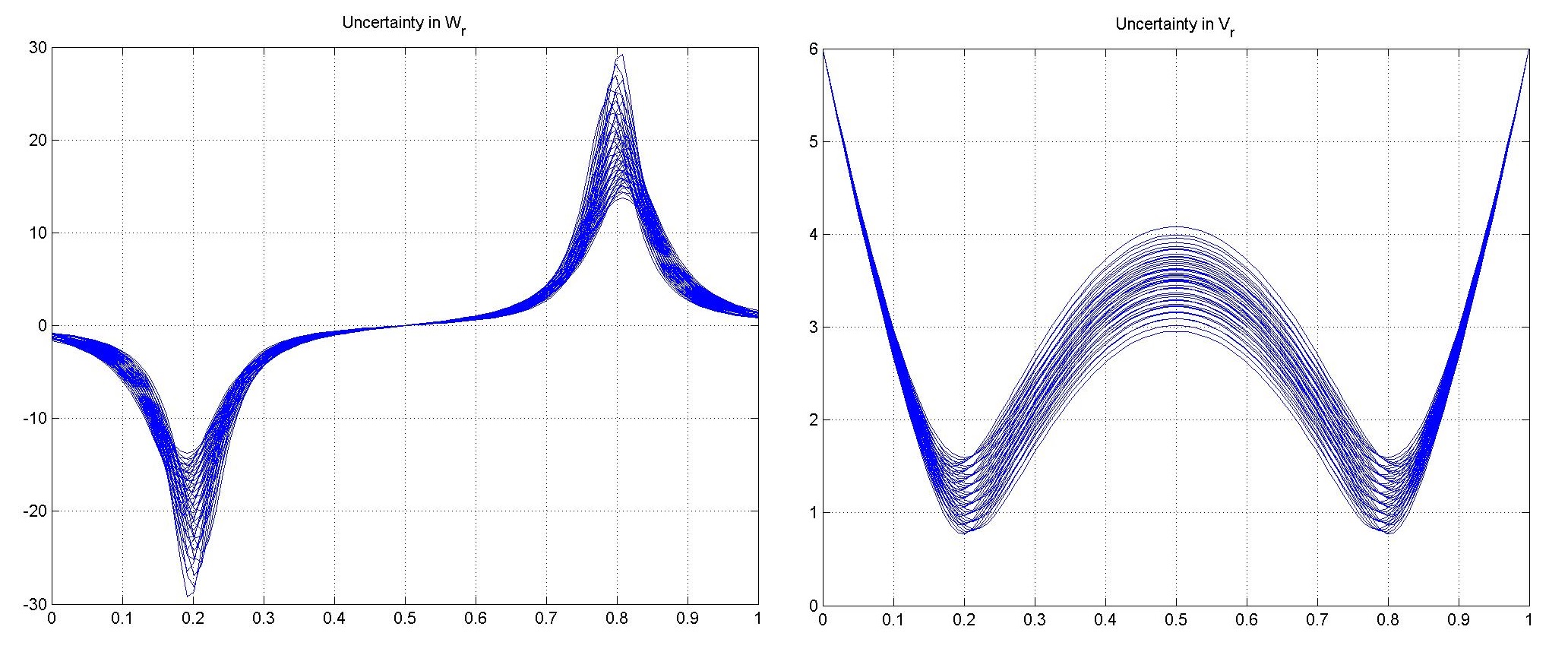}
    \caption{Uncertainty profiles for changes in $x_i$ and $y_i$.}
    \label{fig: Uncertainty_xy}
\end{figure*}

Therefore, the parameters $(x_i,y_i )$, which include the above profiles, are the original uncertainty parameters of the control system. We calculate feedback-state gains in case of changes in the robot's initial position so that the controller can regulate the system perfectly and the close loop system is robust to these changes. By considering space of uncertainty as $x_i=[0.8 1.1]$ and $y_i=[0.6 0.8]$ the profiles of uncertainties are as shown in Fig. \ref{fig: Uncertainty_xy}.

\begin{figure*}[h]
    \centering
    \includegraphics[width=\textwidth]{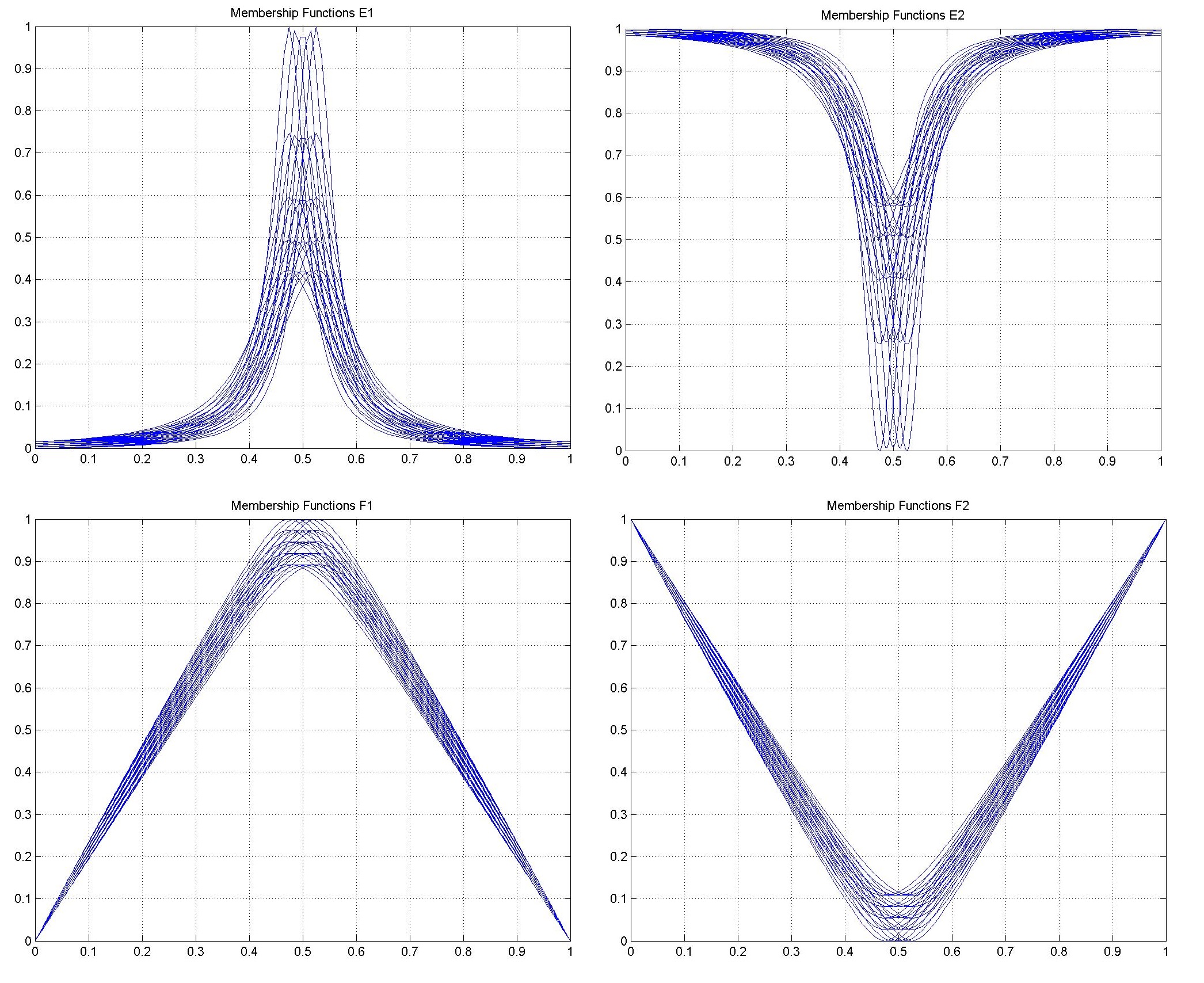}
    \caption{Membership function of E and F.}
    \label{fig: membership_2_EF}
\end{figure*}

\begin{figure*}[h]
    \centering
    \includegraphics[width=\textwidth]{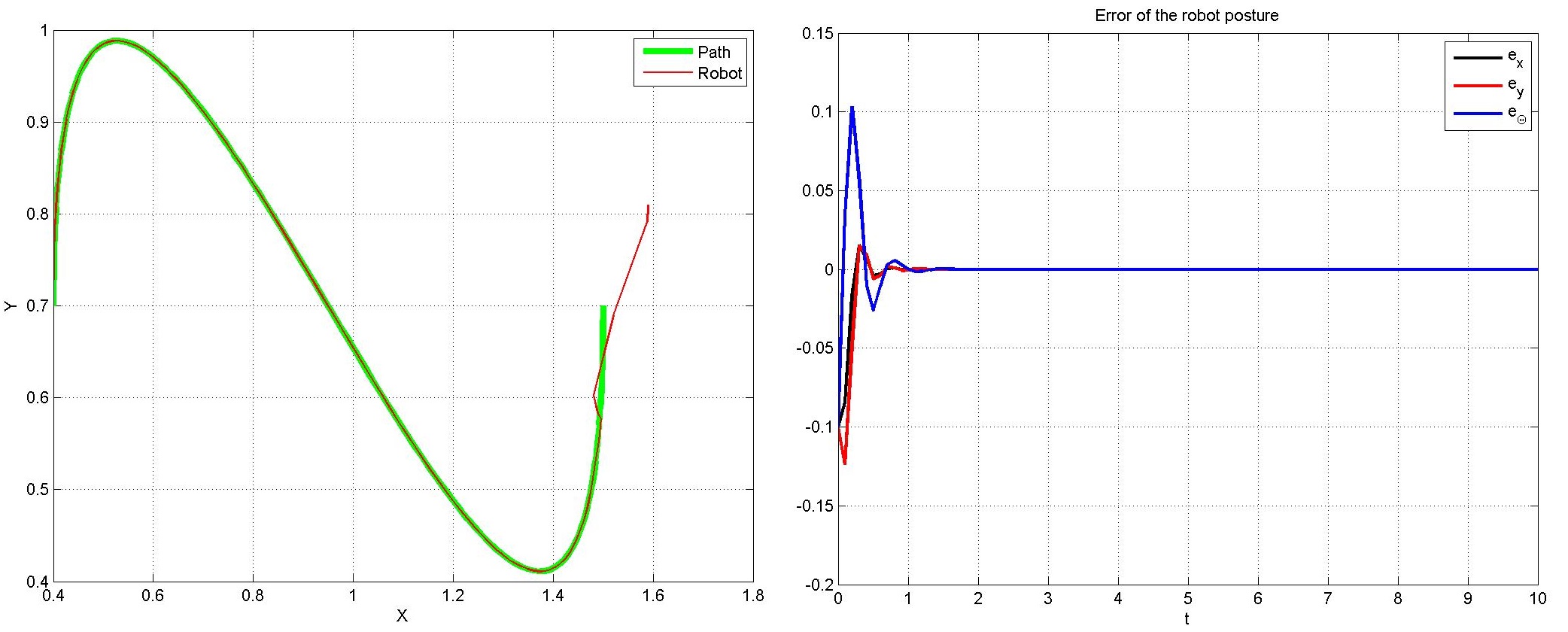}
    \caption{WMR with Type I Fuzzy Controller.}
    \label{fig: type_1}
\end{figure*}

\begin{figure*}[h]
    \centering
    \includegraphics[width=14cm, height = 10cm]{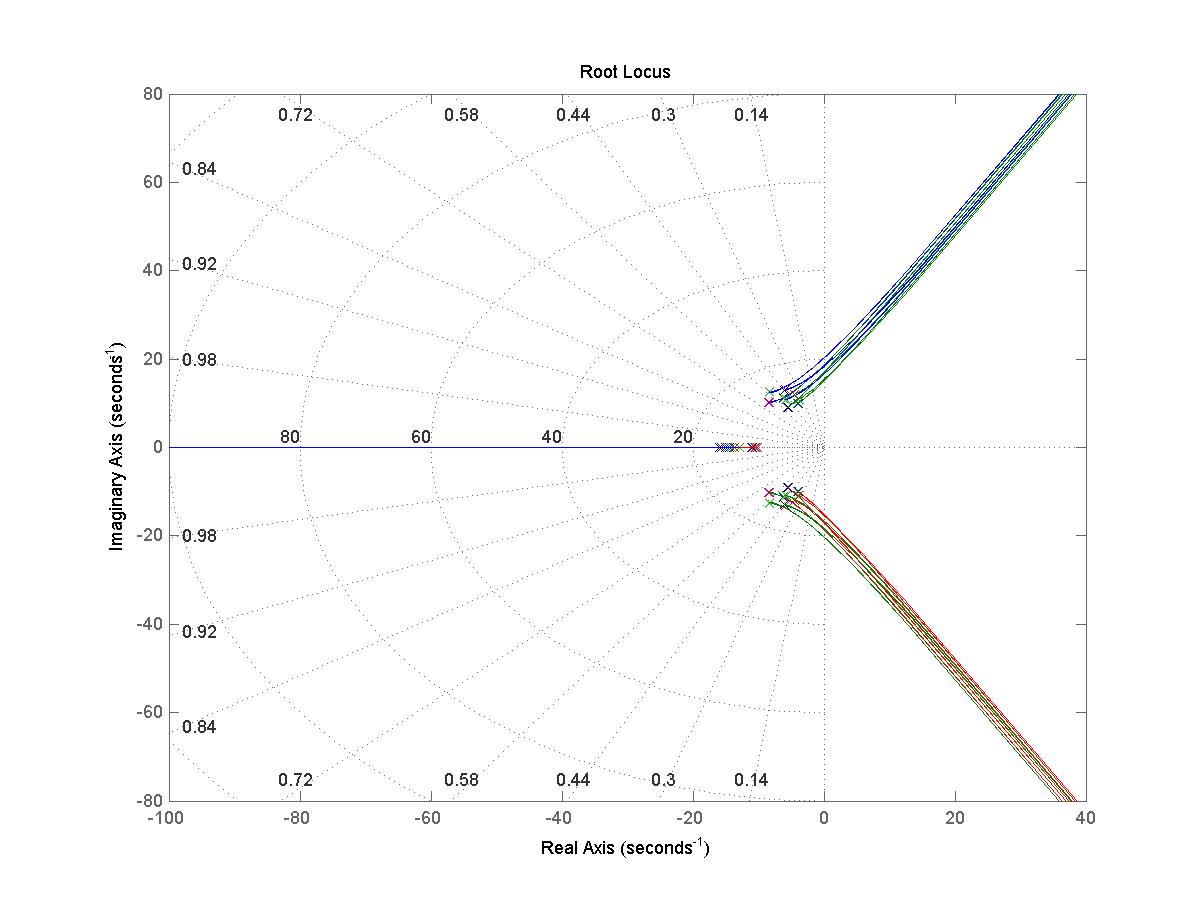}
    \caption{Location of Roots for Type I Fuzzy Controller.}
    \label{fig:root_location_1}
\end{figure*}

\begin{figure*}[h]
    \centering
    \includegraphics[width=\textwidth]{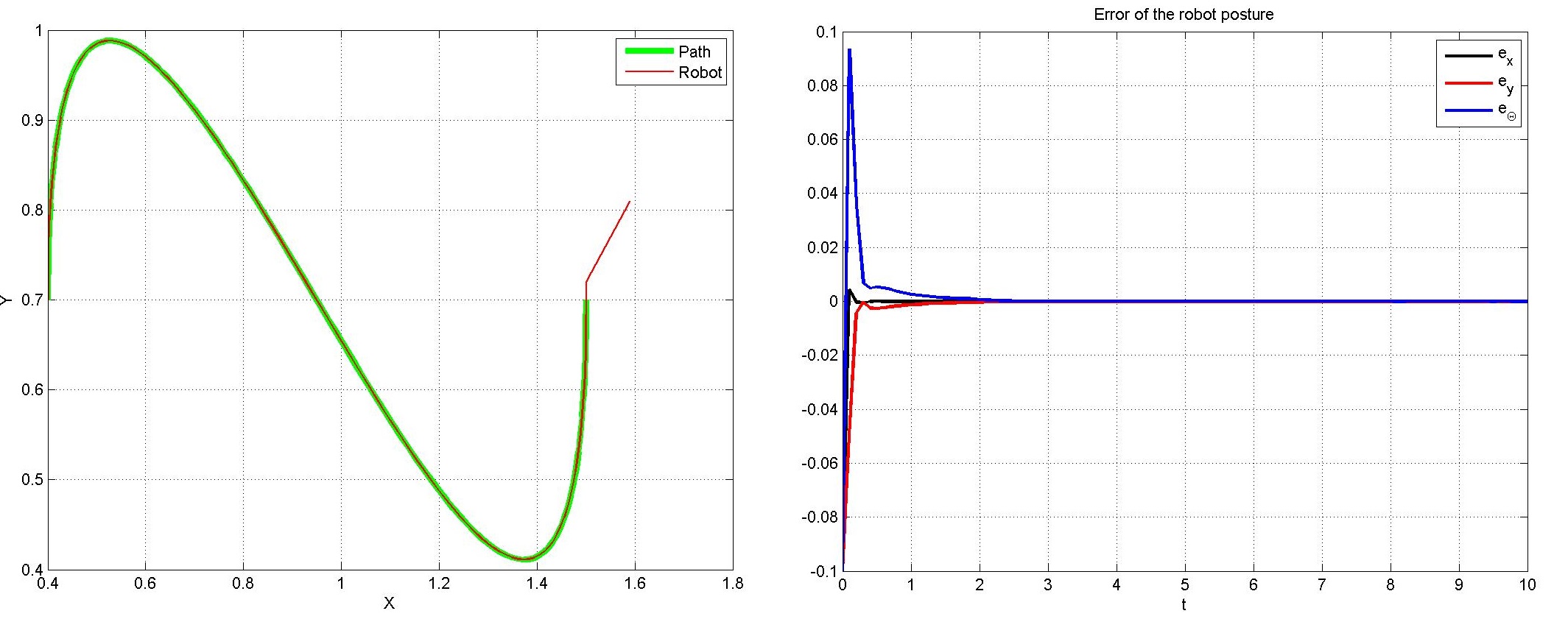}
    \caption{WMR with Type II Fuzzy Controller.}
    \label{fig: type2}
\end{figure*}

For eq. (\ref{eq:main}) and the same conditions of  eq. (\ref{eq:zoning}) and (\ref{eq:error}), we consider the same linguistic variables of Type I for Type II. Therefore
\begin{equation} \label{eq:eqa}
    \begin{bmatrix}\dot e_x \\  \dot e_y \\ \dot e_\theta\end{bmatrix}=
    \begin{bmatrix} 0 & z_1 & 0\\ -z_2 & 0 & z_2\\ 0 & 0 & 1 \end{bmatrix}
    \begin{bmatrix} e_x \\ e_y \\ e_\theta\end{bmatrix}-
    \begin{bmatrix} 1 & -z3\\  0 & z_4\\ 0 & 1 \end{bmatrix} U_B
\end{equation}
Notably, the max and min of each linguistic variable are calculated based on all states and considering the uncertainties in the initial point of WMR. Viewing the below profile of uncertainties, Figures \ref{fig: uncertainty_theta} and \ref{fig: Uncertainty_xy}, with the same zoning of Type I, led to the membership function of the Type II controller.
\begin{equation}
    \begin{cases}
        x_i \in [x_{i,1},x_{i,N}] \\
        y_i \in [y_{i,1},y_{i,N}]
    \end{cases}
\end{equation}
Therefore: 
\begin{equation}
  \begin{cases}
    x_{i,j}=x_{i,1}+j\frac{x_{i,N}-x_{i,1}}{N}\quad N \in \mathbb{N} \\
y_{i,j}=y_{i,1}+j\frac{y_{i,N}-y_{i,1}}{N}\quad N \in \mathbb{N}
  \end{cases}
\end{equation}
these uncertainties affect the maximum and the minimum velocities used in zoning eq. \ref{eq:minmax}.
Therefore, in the defined range:
\begin{equation}
z_1(t)=\sum_{i=1}^{2} \widetilde{E}_i (z_1(t))e_i
\end{equation}
while
\begin{equation}
\widetilde{E}_1(z_1(t))|_{x_{i,j},y_{i,j}}+\widetilde{E}_2(z_1(t))|_{x_{i,j},y_{i,j}}=1
\end{equation}
Therefore:
\begin{equation}
  \begin{cases}
    \widetilde{E}_1(z_1(t))=\frac{z_1(t)-e_2}{e_1-e_2}\\
    \widetilde{E}_2(z_1(t))=\frac{e_1-z_1(t)}{e_1-e_2}
    \end{cases}
\end{equation}
and in same way for $z_2,z_3 $ and $z_4$. Fig. \ref{fig: membership_2_EF} shows that the first and second linguistic variables are still Type I, and the third and fourth are Type II.


The IF-THEN rule box base on the \underline{T-S model} is:
Rule $i^{th}$ of the ``IF-THEN" rules set:
 \[if \quad z_1 (t)\quad is \quad\widetilde{M_{i1}} \quad and \quad \cdots \quad and \quad z_p (t)\quad is \quad \widetilde{M_{ip}}\]
\begin{equation}
then \begin{cases}
    \dot x(t)=A_ix(t)+B_iu(t)\\
    y(t)=C_ix(t)
    \end{cases} i=1,\cdots\,r
\end{equation}
The above definition implies the only difference between Type I and Type II is in the $M_{ip}$ expressions which are type II membership functions. Still, we will see further on that this increase in dimension will lead to a rise in the equations in a staggering manner. Considering up and down boundaries in each $M_{ip}$ that the whole system operates in it, we have for the lower bound:
\begin{equation}
\underline{\dot x}(t)=\sum_{i=1}^{r} \underline{h}_i(Z(t)){A_ix(t)+B_iu(t)}
\end{equation}
\begin{equation}
\underline{\dot y}(t)=\sum_{i=1}^{r} \underline{h}_i(Z(t))C_ix(t)
\end{equation}
where
\begin{equation}
\underline{W}_i(Z(t))=\prod_{j=1}^{p}\underline{M}_{ij}z_j(t)
\end{equation}
\begin{equation}
\underline{h}_i(Z(t))=\frac{\underline{W}_i(Z(t))}{\sum_{i=1}^{r}\underline{W}_i(Z(t))}
\end{equation}
And for the upper bound:
\begin{equation}
\overline{\dot x}(t)=\sum_{i=1}^{r} \overline{h}_i(Z(t)){A_ix(t)+B_iu(t)}
\end{equation}
\begin{equation}
\overline{\dot y}(t)=\sum_{i=1}^{r} \overline{h}_i(Z(t))C_ix(t)
\end{equation}
where
\begin{equation}
\overline{W}_i(Z(t))=\prod_{j=1}^{p}\overline{M}_{ij}z_j(t)
\end{equation}
\begin{equation}
\overline{h}_i(Z(t))=\frac{\overline{W}_i(Z(t))}{\sum_{i=1}^{r}\overline{W}_i(Z(t))}
\end{equation}
The overall output of the controller is: 
\begin{equation} \label{eq:output}
y(t)=\frac{\sum_{i=1}^{r} \underline{h}_i(Z(t))C_ix(t)+\sum_{i=1}^{r} \overline{h}_i(Z(t))C_ix(t)}{2}
\end{equation}

Eq. (\ref{eq:output}) shows that Type II has up and down boundaries, that each one is a Type I. Base on PDC for the corresponding rule, we have:
\begin{equation}
u_B(t)=-f_i x(t)
\end{equation}
Therefore, the PDC can lead to follow relations:
\begin{equation}
\underline{u}_B(t)=-\frac{\sum_{i=1}^{r}\underline{W}_i(Z(t))F_ix(t)}{\sum_{i=1}^{r}\underline{W}_i(Z(t))}=-\sum_{i=1}^{r}\underline{h}_i(Z(t))F_ix(t)
\end{equation}
\begin{equation}
\overline{u}_B(t)=-\frac{\sum_{i=1}^{r}\overline{W}_i(Z(t))F_ix(t)}{\sum_{i=1}^{r}\overline{W}_i(Z(t))}=-\sum_{i=1}^{r}\overline{h}_i(Z(t))F_ix(t)
\end{equation}
which
\begin{equation}
u_B(t)=-\frac{\sum_{i=1}^{r} \underline{h}_i(Z(t))C_ix(t) + \sum_{i=1}^{r} Z(t)C_ix(t) \overline{h}_i}{2}
\end{equation}
And in the last point, the combination of eq. (\ref{eq:PDC1}) and (\ref{eq:eqa}) makes the overall system as follows:
\[\dot x(t)=\frac{1}{2} \sum_{i=1}^{r} \overline{h}_i(Z(t))\overline{h}_i(Z(t)G_{ii}x(t)\]
\[+\sum_{i=1}^{r}\sum_{i<j}^{}\overline{h}_i(Z(t))\overline{h}_j (Z(t))\frac{G_{ij}+G_{ji}}{2}x(t)\]
\[+\frac{1}{2} \sum_{i=1}^{r}\underline{h}_i (Z(t))\underline{h}_i (Z(t))G_{ii} x(t)+\]
\begin{equation}
\sum_{i=1}^{r}\sum_{i<j}^{}\underline{h}_i (Z(t))\underline{h}_j (Z(t))\frac{G_{ij}+G_{ji}}{2}x(t)
\end{equation}
where,
\begin{equation}
G_{ij}=A_i-B_i F_j
\end{equation}

As for type I, we must find a positive defined matrix $P$ that satisfies the Lyapunov conditions. The LMI problem changes to the below shape:
Finding the values of $X>0$ and $M_i$ that meet the following relations:
\[-XA_i^T-A_i X+M_i^T B_i^T+B_i M_i>0\]
\[-XA_i^T-A_i X-X A_j^T-A_j X + M_j^T B_i^T+\]
\[B_i M_j+M_i^T B_j^T+B_j M_i \geq 0\]
\[i<j \]
for \begin{equation}
\underline{h}_i \cap \underline{h}_j \neq \varnothing \quad and \quad \overline{h}_i \cap \overline{h}_j \neq \varnothing 
\end{equation}
While 
\begin{equation}
M_i=F_i X \quad and \quad X=P^{-1}
\end{equation} 

\section{Evaluation} \label{sec:evaluation} \subsection{Key Findings}

Two types of controllers, Type I and Type II, were considered for the designed fuzzy controllers. Each controller's performance is evaluated by studying the convergence of the system errors and the positions of the closed-loop poles. For Type I fuzzy controller, the initial error $e_0=\begin{bmatrix}-0.1&-0.1&-6\end{bmatrix}^T$ for the path shown in Fig. \ref{fig:Paths}.a was considered. The iteration $98^{th}$ result of the fuzzy control rules produced a control gain matrix $P$ of:
\begin{equation}
    P=\begin{bmatrix}
    0.5756 &0.0102& 0.0139\\0.0102&0.9640 & 0.6352\\0.0139&0.6352&0.6690 \end{bmatrix}
\end{equation}
and the local controller matrices are:
    \[F_1=\begin{bmatrix}
     -13.4139   & 2.8816 &  -0.2719\\ -0.5201 & -14.1938 & -15.2651 \end{bmatrix}\]
     \[\vdots\]
\begin{equation}
    F_{16}=\begin{bmatrix}
     -12.5896 &  -4.2942  & -0.2809\\0.3038 &  -10.7937  & -14.8515)\end{bmatrix}     
\end{equation}

The local controller matrices $F i$ for every parallel compensating controller are obtained. Fig. \ref{fig: type_1} displays the errors for the Type I controller and highlights how the error decreases much more quickly than with the classic controller (Fig. \ref{fig:classic}). The WMR was also capable of continuing its path despite facing fluctuations at the path's entrance.

Fig. \ref{fig:root_location_1} shows the locations of the closed-loop poles. In all cases, and in some cases up to three times greater than the real component of the imaginary roots, it is noted that the real poles change between 10.17 and 15.61. Also, the zeta value ranges between 0.34 and 0.64, which is suitable for a practical system. So, it can be said that the designed Type I fuzzy controller stabilizes the WMR satisfactorily and performs better than the traditional controller.

For Type II fuzzy controller, the same initial error $e_0=\begin{bmatrix}-0.1&-0.1&-6\end{bmatrix}^T$ for the 
path in Fig. \ref{fig:Paths}.a was considered, and the fuzzy control rules generated a control gain matrix $P$ of:
\begin{equation}
P= \begin{bmatrix}5.5241  &  0.0212  &  0.0127\\0.0212  &  6.2709 &   2.0279\\0.0127  &  2.0279 &   1.1183]
    \end{bmatrix}
\end{equation}
as well as $F i$ local controller matrices. Fig. \ref{fig: type2} displays the errors for the Type II controller, illustrating how the WMR moves more smoothly and the fluctuation is eliminated. The local controller matrices are:
\[F_1=\begin{bmatrix}-38.2344 &  10.7068 &  -1.8843\\ 2.7505 & -35.2521 & -19.5316\end{bmatrix}\]
\[\vdots\]
\begin{equation}
F_{16}=\begin{bmatrix}-32.9008 & -17.0932  &  0.4884\\-1.9632 & -32.1240 & -19.8303\end{bmatrix}
\end{equation}
It is important to note that the suggested controllers can handle parameter uncertainty within the specified ranges. It may not be possible to consider any arbitrary interval as the uncertainty of parameters because the LMI problem may not have an answer. Accordingly, the uncertainty interval can be considered as follows:
\begin{equation} \label{eq:points}
    \begin{cases}
        x_i  \in [0.7,1.5] \\ y_i \in [0.5,0.9] 
    \end{cases}
\end{equation}

\begin{figure*}[h]
    \centering
    \includegraphics[width=14cm, height = 10cm]{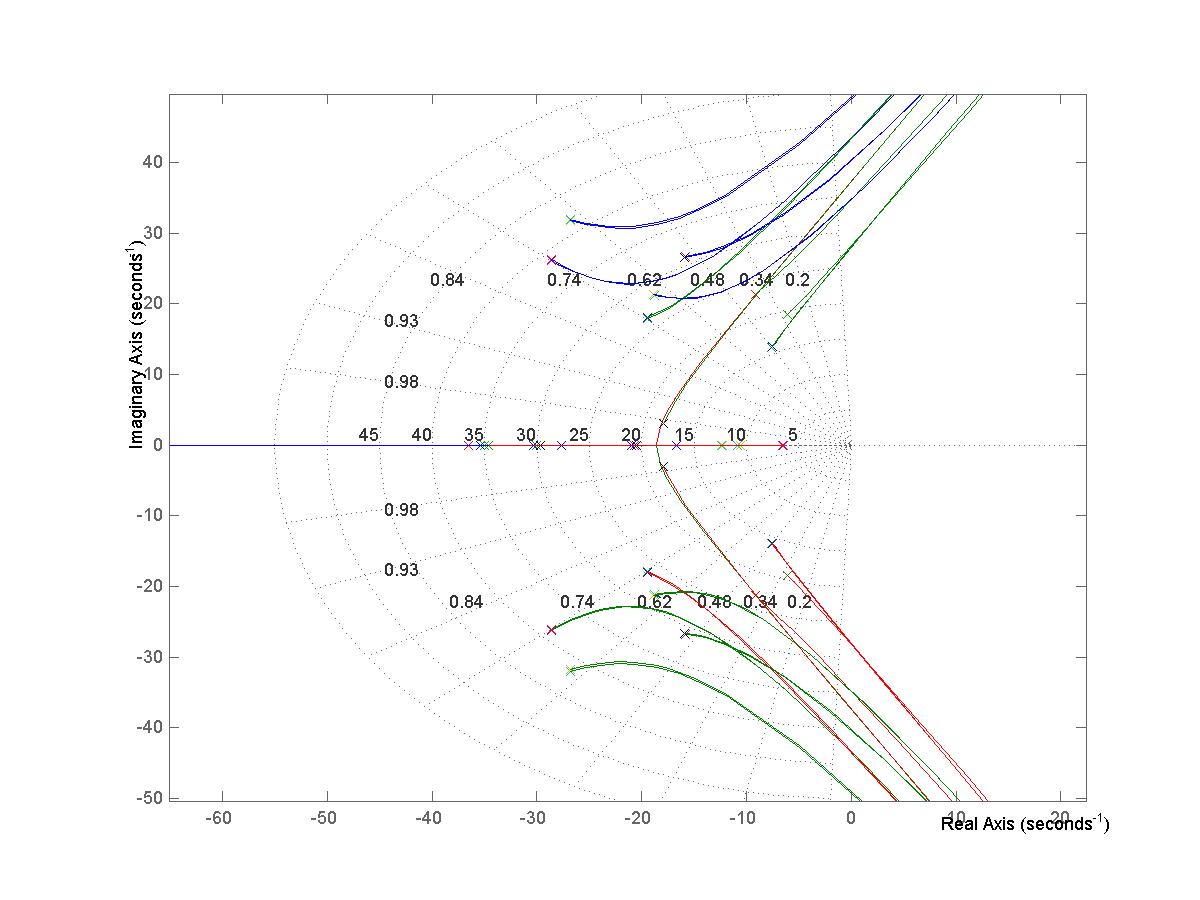}
    \caption{Location of Roots for Type II Fuzzy Controller.}
    \label{fig: root_location_2}
\end{figure*}

Fig. \ref{fig: root_location_2} shows the locations of the closed-loop poles for the Type II fuzzy controller. It has been determined that the real pole portions begin at -6.49 and end at -36.6. The imaginary poles are also located on the left side indicating that the controller is stable at any point along the path. As a result, it can be said that the proposed Type II fuzzy controller achieves smoother and more stable performance than the Type I controller and successfully maintains the WMR.

Because of their ability to manage the WMR system's uncertainties, the designed fuzzy controllers are usable in real-world situations.

\subsection{Future Directions}
Last, while these controllers have shown promising results in various applications, there are several potential future directions to explore in order to enhance their capabilities. Such directions include the following.
\begin{itemize}
    \item Adaptive Fuzzy Controllers: Incorporate adaptive mechanisms into fuzzy controllers to dynamically adjust the fuzzy sets, rules, and control parameters based on the robot's operating conditions and environment. Adaptive fuzzy controllers can improve the performance and robustness of WMRs by continuously adapting to changing conditions.
    \item Learning and Optimization Techniques: Integrate machine learning and optimization techniques with fuzzy controllers to enhance their performance. For example, evolutionary algorithms, reinforcement learning, or neural networks can be used to optimize the fuzzy controller's rule base or tune the membership functions and control parameters.
    \item Hybrid Control Approaches: Combine fuzzy control with other control strategies, such as model predictive control, sliding mode control, or adaptive control, to leverage the strengths of each approach. Hybrid control systems can provide better performance, stability, and robustness compared to individual controllers.
    \item Cooperative and Multi-Robot Systems: Extend fuzzy control approaches to cooperative and multi-robot systems. Develop coordination and cooperation strategies among multiple WMRs using fuzzy controllers to achieve tasks that require collaboration, such as swarm robotics, formation control, or object manipulation.
\end{itemize}


\section{Conclusion} This paper presents a path following controller for a wheeled mobile robot or WMR in a third-order Bezier curve. The Bezier curve can generate different paths. The controller architecture includes both feedforward and feedback controllers, and we used a classical controller approach to linearize the system around the operating point. We have explored the application of fuzzy type I and type II logic in a particular domain and demonstrated its effectiveness aligns with classic strategies through a case study. By dividing the controller into smaller parts, analyzing each rule separately, and mapping the rules into parallel distributed compensation using LMI analysis, the proposed parallel distributed compensation, sector of nonlinearity, and local approximation approach can address the complexity and if-then issue of fuzzy controllers. This method reduces uncertainty in the equations, avoids the curse of dimensionality, and simplifies the controller by approximating nonlinear functions with simpler ones in different input areas. These strategies can potentially improve the resilience and efficiency of the fuzzy controller. The proposed controller follows the desired path with minimum position error and in a smooth motion. Our findings suggest that fuzzy type I and type II logic can lead to more accurate and robust models in situations with uncertainty and imprecision.

 \label{sec:conclusion}

\section*{Acknowledgment}
The authors would like to acknowledge Professor Faridoon Shabaninia for their profound knowledge that has been instrumental in shaping the content and direction of this paper.

\bibliographystyle{IEEEtran}
\bibliography{references.bib}

\end{document}